\documentclass[runningheads]{llncs}
\usepackage{graphicx}
\usepackage{amsmath}
\usepackage{array}
\newcolumntype{M}[1]{>{\centering\arraybackslash}p{#1}}

\begin{document}
\title{On-Device Language Identification of Text in Images using Diacritic Characters}
%
%
\author{Shubham Vatsal\inst{} \and
Nikhil Arora\inst{} \and
Gopi Ramena\inst{} \and
Sukumar Moharana\inst{} \and
Dhruval Jain\inst{} \and
Naresh Purre\inst{} \and
Rachit S Munjal\inst{}}

%
\authorrunning{}
%
\institute{On-Device AI, Samsung R \& D Institute, Bangalore, India
\centerline{\email{\{shubham.v30,n.arora,gopi.ramena,msukumar,dhruval.jain,naresh.purre,rachit.m\}@samsung.com}}}
\maketitle              
\begin{abstract}
Diacritic characters can be considered as a unique set of characters providing us with adequate and significant clue in identifying a given language with considerably high accuracy. Diacritics, though associated with phonetics often serve as a distinguishing feature for many languages especially the ones with a Latin script. In this proposed work, we aim to identify language of text in images using the presence of diacritic characters in order to improve Optical Character Recognition (OCR) performance in any given automated environment. We showcase our work across 13 Latin languages encompassing 85 diacritic characters. We use an architecture similar to Squeezedet for object detection of diacritic characters followed by a shallow network to finally identify the language. OCR systems when accompanied with identified language parameter tends to produce better results than sole deployment of OCR systems. The discussed work apart from guaranteeing an improvement in OCR results also takes on-device (mobile phone) constraints into consideration in terms of model size and inference time.

\end{abstract}
\section{Introduction}

A diacritic or diacritical mark is basically a glyph added to a letter or a character. Diacritics are used to provide extra phonetic details and hence altering the normal pronunciation of a given character. In orthography \footnote{https://en.wikipedia.org/wiki/Orthography}, a character modified by a diacritical mark is either treated as a new character or as a character-diacritic combination. These rules vary across inter-language and intra-language peripherals. In this proposed work, we have restricted ourselves to diacritic characters pertaining to Latin languages. Other than English there are many popular Latin languages which make use of diacritic characters like Italian, French, Spanish, German and many more.

OCR is one of the most renowned and foremost discussed Computer Vision (CV) tasks which is used to convert text in images to electronic form in order to analyze digitized data. There have been many prominent previous works done in OCR. \cite{wojna2017attention} uses a novel mechanism of attention to achieve state of the art results on street view image datasets. \cite{bartz2017see} makes use of spatial transformer network to give unparalleled results in scene text recognition. \cite{smith2016end} applies conventional Convolutional Neural Network (CNN) with Long Short Term Memory (LSTM) for its text interpretation task.

We can define two broad ways with respect to OCR enhancements. One can be an implicit way of OCR enhancement whereas other can be an explicit way. In the explicit way of OCR enhancement our aim is to improve OCR's inherent accuracy which can depend on multiple factors like OCR's internal architecture, pre-processing images to improve their quality and hence increasing OCR's relative confidence with regards to text recognition and so on. The quality of image depends on multiple aspects with respect to OCR performance ranging from font size of text in images to source of images. There are many image pre-processing techniques like \cite{bieniecki2007image} \cite{lat2018enhancing} \cite{seeger2001binarising} which help in enhancing image quality and in return provide us with better OCR confidence. The other type of OCR enhancements are the implicit ones. In this way of OCR enhancement, we concentrate on external factors in order to improve OCR's results in a mechanized environment. For example, post processing hacks to improve OCR results, determining factors like language of text in image and using them as OCR parameters to choose the correct OCR language based dependencies are some of such factors. An important point to emphasize here is that an OCR's original accuracy stays the same in case of implicit enhancements but the final OCR results in a given environment is improved. In this work we concentrate on one of the implicit ways to improve OCR results. Language input to OCR helps in differentiating between similar looking characters across various languages which comprise mostly of diacritic characters. For example, diacritic characters \emph{\`{a}} and \emph{\'{a}} are minutely different and hence if correct language is not specified, it is often missed or wrongly recognized.

The rest of the paper is organised in the following way. Section 2 talks about related works. We elucidate the working of our pipeline in section 3. Section 4 concentrates on the experiments we conducted and the corresponding results we achieved. The final section takes into consideration the future improvements which can be further incorporated.

\section{Related Works}

There have been many works done to identify languages in Natural Language Processing (NLP) domain but things are not that straightforward when it comes to identifying languages of text in images, especially when it needs to be done without any involvement of character segmentation or OCR techniques. Most of the existing works on OCR implicitly assume that the language of the text in images is known beforehand. But, OCR approaches work well individually for specific languages for which they were designed in the first place. For example, an English OCR will work very well with images containing English text but they struggle when given a French text image. An automated ecosystem would clearly need human intervention in order to select the correct OCR language parameters. A pre-OCR language identification work would allow the correct language based OCR paradigms to be selected thus guaranteeing better image processing. Along the similar lines, when dealing with Latin languages, current OCR implementations face problems in correct classification of languages particularly due to common script. In this paper, we propose an architecture which uses detection of diacritic characters in all such languages using object detection approach to enhance the OCR text recognition performance. Key takeaway from our approach is that we design this pipeline to meet the on-device constraints, making it computationally inexpensive.

Several work has been done with respect to script detection but identification of language from images is still not a thoroughly researched area. Script detection although could help us in differentiating two languages of different scripts but this technique fails to differentiate between languages of same script like Spanish and German which belong to Latin script. Among some of the previous works done in the domain of language identification, \cite{elgammal2001techniques} uses three techniques associated with horizontal projection profiles as well as runlength histograms to address the language identification problem on the word level and on text level. But then this paper just targets two languages which are English and Arabic who also happen to have different scripts. \cite{peake1997script} although with the similar intention of improving OCR showcases its work only on languages of different scripts. Again, \cite{zhu2008unconstrained} presents a new approach using a shape codebook to identify language in document images but it doesn't explicitly targets languages of similar script. \cite{nicolaou2016visual} demonstrates promising results but then the authors attribute these results towards biased image properties as all texts were of the same size and acquired under exactly the same conditions. \cite{mioulet2015language} advocates that the use of shape features for script detection is efficient, but using the same for segregating into languages is of little importance as many of these languages have same set of characters. Also this work uses an OCR for identification of language contrary to our work where we aim to identify language first and then use it to improve OCR.

Some noteworthy works revolving around diacritic character in images include robust character segmentation algorithm for printed Arabic text with diacritics based on the contour extraction technique in \cite{mohammad2019contour}. Furthermore, diacritic characters have been used for detecting image similarity in Quranic verses in \cite{alotaibi2017optical}. Another work \cite{gajoui2015diacritical} discusses about diacritical language OCR and studies its behaviours with respect to conventional OCR. \cite{majid2019segmentation} talks about their segmentation-free approach where the characters and associated diacritics are detected separately with different networks. Finally,\cite{lutf2014arabic} illustrates experiments on Arabic font recognition based on diacritic features. None of these works try to associate diacritic characters with language as we have explored in our case.

Object Detection is a widely popular concept which has seen many breakthrough works in the form of Fast R-CNN \cite{girshick2015fast}, YOLO \cite{redmon2016you}, SqueezeNet \cite{iandola2016squeezenet} and many more. There have been quite a few works in the direction of using object detection approach for character recognition. \cite{wang2011end} uses a generic object recognition technique for end to end text identification and shows how it performs better than conventional OCR. \cite{li2020occluded} makes use of deep convolutional generative adversarial network and improved GoogLeNet to recognise handwritten Chinese characters. In our work also, we make use of object detection mechanism with Squeezedet to process diacritic characters.

Other previous approaches on OCR for Latin language identification fail to perform well after script detection phase. To the best of our knowledge diacritic characters have not been used for the same to enhance the system performance. In this paper, we present a novel architecture for boosting OCR results when it comes to working with different languages with common scripts, with an efficient performance when deployed on-device.

\section{Proposed Pipeline}
\begin{figure}
\centerline{\includegraphics[width=1.0\linewidth]{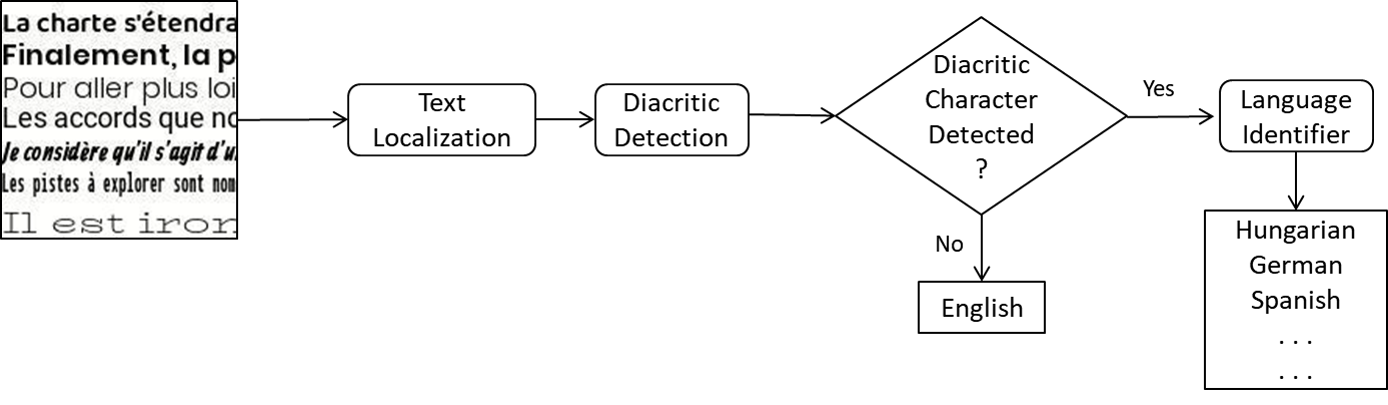}}
\caption{Proposed Pipeline}
\label{fig:pipeline}
\end{figure}

This section delineates the purpose of each component and eventually concludes how these components blend together to get us the desired result. Fig. \ref{fig:pipeline} shows the pipeline of the proposed system. As we can see, an image is sent as input to a Text Localization component from which text bounding boxes are extracted.  These text bounding boxes are sent one by one to Diacritic Detection model. Once the diacritics if present have been detected, then we use our shallow neural network to identify the language. This language input is finally fed to the OCR to improve its performance.

\subsection{Corpus Generation}
\label{cg}
\begin{figure}
\centerline{\includegraphics[width=1.0\linewidth]{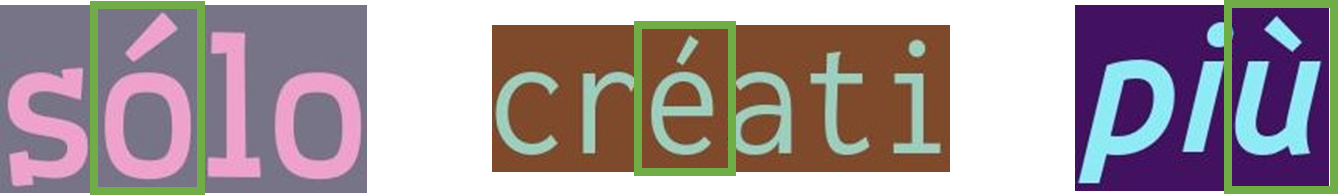}}
\caption{Sample Images}
\label{fig:corpus}
\end{figure}

We created RGB format word image dataset of fixed height of 16 dimension and variable width depending on the aspect ratio to train our model for diacritic characters. We used European Parliament Proceedings Parallel Corpus \footnote{http://www.statmt.org/europarl/} for purposefully choosing words with diacritic characters across all 13 languages for constructing this dataset. The distribution of data across all languages and the diacritic characters found in each language is listed in Table \ref{tab:Corpus Distribution}. We uniquely labelled each diacritic character. In order to achieve an adequate level of generalization, various randomization factors were put into place like font size, font type and word length. Sample snippets of this synthetic dataset have been showcased in Fig. \ref{fig:corpus}. As it can be seen in the figure, bounding boxes have been constructed around the diacritic characters.

\begin{table}
\label{tab:Corpus Distribution}
\centering
\caption{Corpus Distribution}\label{tab1}
\begin{tabular}{| M{1.5cm} | M{2.1cm} | M{8.4cm} |}
\hline
\bfseries {Language} & \bfseries{Word Image Corpus Size} & \bfseries{Diacritic Characters}\\
\hline
Spanish & 9218 &  Á, á, Ñ, ñ \\
German & 8673 &  Ä, ä, Ö, ö, Ü, ü, ß \\
French & 9127 &  À, à, Â, â, É, é, È, è, Ê, ê, Ë, ë, Î, î, Ï, ï, Ô, ô, Œ, œ, Û, û, ç \\
Italian & 8903 &  À, à, Ì, ì, Ò, ò, Ù, ù \\
Romanian & 9583 &  Â, â, Ă, ă, Ş, ş, Ţ, ţ \\
Finnish & 9477 &  Ä, ä, Ö, ö \\
Hungarian & 9674 &  Á, á, É, é, Í, í, Ó, ó, Ö, ö, Ő, ő, Ü, ü, Ű, ű \\
Estonian & 9243 &  Ä, ä, Õ, õ, Ö, ö, Š, š \\
Danish & 9251 & Å, å, Æ, æ, Ø, ø \\
Dutch & 9439  & Ë, ë, Ï, ï \\
Swedish & 9054 &  Ä, ä, Å, å, Ö, ö \\
Portuguese & 8891 &  Á, á, Ã, ã, Ê, ê, Ô, ô, Õ, õ, ç \\
Czech & 9133 &  Á, á, É, é, Ě, ě, Í, í, Ó, ó, Ú, ú, ů, Ý, ý, Č, č, Ď, ď, Ň, ň, Ř, ř, Š, š, Ť, ť, Ž, ž \\

\hline
\end{tabular}
\label{tab:Corpus Distribution}
\end{table}

Apart from the above discussed word based image dataset we also created RGB format 150x150 Test dataset. This dataset was again created using European Parliament Proceedings Parallel Corpus in order to test the final performance of our proposed pipeline. This dataset consisted of random text containing some diacritic characters which was fed as an input to our pipeline. We again took care of same set of randomization factors in order to achieve a better level of generalization. Sample image of this dataset can be seen in Fig. \ref{fig:pipeline}.
\begin{figure}
\centerline{\includegraphics[width=1.0\linewidth]{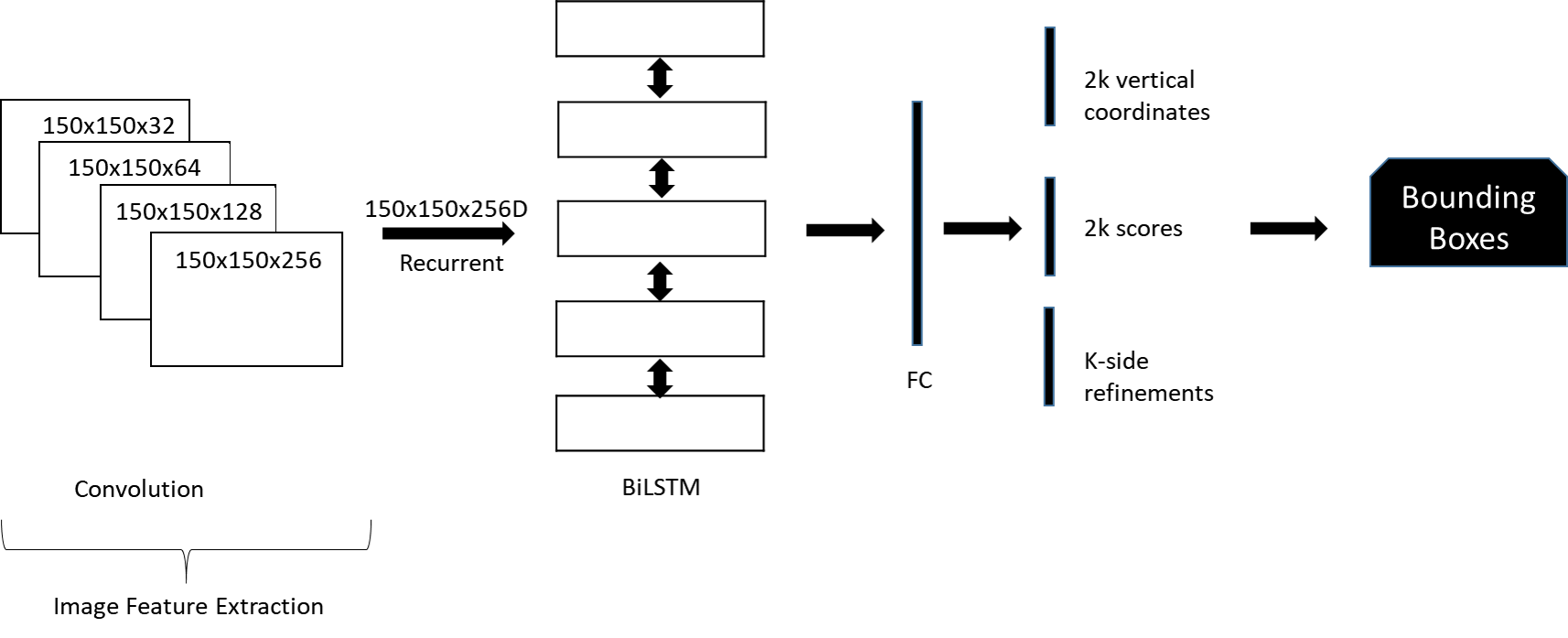}}
\caption{Text Localization (Modified CTPN Architecture)}
\label{fig:tl}
\end{figure}

\subsection{Text Localization}


Text localization detects bounding boxes of text regions. This is performed using Connectionist Text Proposal Network (CTPN) \cite{tian2016detecting}. We modified the network to use a 4 layered CNN instead of VGG 16 \cite{simonyan2014very}, to achieve a better on-device performance and also since we needed only low level features in order to identify the regions of text. The 4 layers of CNN used similar parameters as that of initial layers of VGG 16 and the filter size of convolutional layers can be seen in Fig. \ref{fig:tl}. Apart from the 4 layered CNN introduced for on-device compatibility, rest of the architecture has been kept same with same parameters as discussed in \cite{tian2016detecting}. The extracted feature vectors are recurrently connected by a Bi-directional LSTM, where the convolutional features are used as input of the 256 dimension Bi-LSTM. This layer is connected to a 512 dimension fully-connected layer, followed by the output layer, which jointly predicts text/non-text scores, y-coordinates and side-refinement offsets of k anchors. The detected text proposals are generated from the anchors having a text/non-text score of $>$ 0.7 (with non-maximum suppression). The modified network architecture of CTPN has been represented in Fig. \ref{fig:tl}. In our experiments, we notice that this is able to handle text lines in a wide range of scales and aspect ratios by using a single-scale image, as mentioned in the original paper.

\subsection{Diacritic Detection}

\begin{figure}
\centerline{\includegraphics[width=1.0\linewidth]{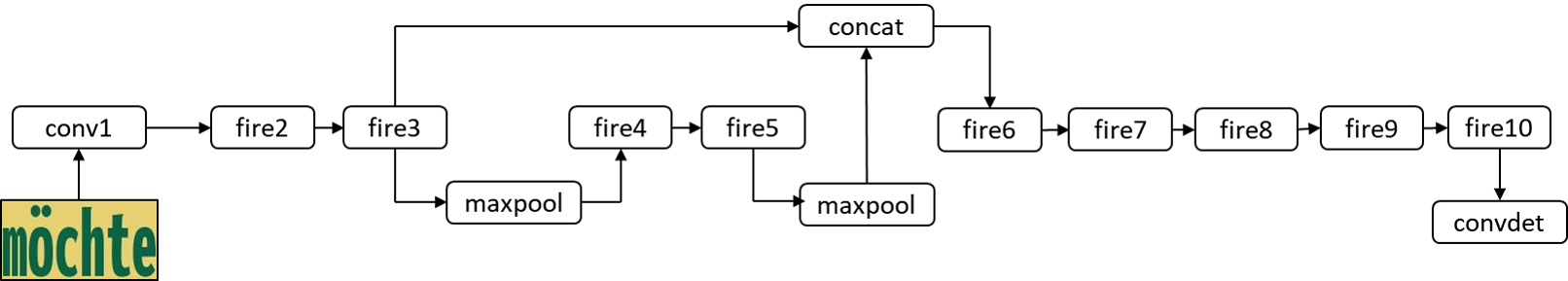}}
\caption{Diacritic Detection Network}
\label{fig:ddn}
\end{figure}

We use an object detection approach to detect diacritic characters. Inspired from Squeezedet \cite{wu2017squeezedet}, we designed a model which is more suitable for our problem statement and also more lightweight in terms of on-device metrics. Since, there are a lot of similarities between normal characters and diacritic characters and also within various diacritic characters, we used our own downsizing network in the initial layers so that sharp difference between various characters could be identified. We didn't use pooling layers in the starting of the network to allow more low level image features to be retained till that point. Further, we decreased the strides of first CNN layer in order to capture more image features. Apart from these changes, we also reduced the number of fire \cite{iandola2016squeezenet} layers. There were couple of reasons for that change. First, our input image is very small and it is not required to have so many squeeze and expand operations and hence make the network very deep as it is the low level image features which mostly contribute towards identifying a difference between a normal character and a diacritic character or even differentiating within the set of diacritic characters.  Second, we also have to adhere to on-device computational constraints. The architecture of our network can be seen in Fig. \ref{fig:ddn}.

For conv1, we used 64 filters with kernel size being 3 and stride 1. Following conv1 we have a set of two fire layers, fire2 and fire3. Both of them have same set of parameters which are $s_{1x1}$=16, $e_{1x1}$=64 and $e_{3x3}$=64 where s represents squeeze convolutions and e represents expand convolutions. Then comes a max pool layer with kernel size 3, stride 2 and same padding. We again have another set of fire layers, fire4 and fire5, having same set of parameters $s_{1x1}$=32, $e_{1x1}$=128 and $e_{3x3}$=128. Max pool follows this set of fire layers with kernel size 3, stride 2 and same padding. We then concatenate the output of these two sets of fire layers and the concatenated output is fed into a new fire layer, fire6. Fire6 and fire7 have $s_{1x1}$=48, $e_{1x1}$=192, $e_{3x3}$=192. Then we have fire8 and with $s_{1x1}$=96, $e_{1x1}$=384, $e_{3x3}$=384. Finally, we have fire9 and fire10 with $s_{1x1}$=96, $e_{1x1}$=384, $e_{3x3}$=384. As it can be seen, we have gradually increased the filters in fire layers from beginning to end of the network. In the end we have convdet layer with kernel size 3 and stride 1. 

In addition to the above discussed model parameters, there were other important hyper-parameters selected to tune the model. While training, we used 9 anchors per grid with batch size of 16. Learning rate was set to 0.01 with decay factor of 0.0001. The non-maximum suppression threshold was set to 0.2 and dropout value was set to 0.5.

\subsection{Language Identification}

We use a shallow network to finally infer the language once diacritic characters have been identified in the given image. We design the input in the form of one-hot vectors corresponding to the total number of diacritic characters with which our Diacritic Detection model was trained.

\begin{figure}
\centerline{\includegraphics[width=0.5\linewidth]{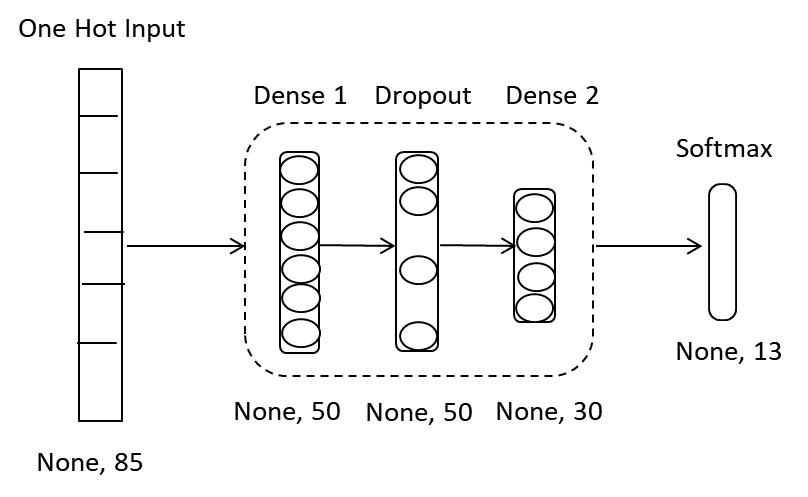}}
\caption{Shallow Network for Language Identification}
\label{fig:shallow}
\end{figure}

We took variable sized chunks of input text and extracted diacritic characters from them to finally prepare our one-hot input vector. Since, we were using European Parliament Proceedings Parallel Corpus for detection of diacritics, we were already having a text dataset labelled on the basis of their language. We used the same dataset to train our shallow network. The shallow network consisted of two hidden dense networks with 50 units and 30 units respectively and ReLu activation function. The output layer consisted of Softmax activation function with number of units being equal to total number of languages which is 13 in our case. The architecture of our network is shown in Fig \ref{fig:shallow}. We created 1000 samples for each language where we used 90\% as training data and remaining as validation data. We trained for 20 epochs with other default parameters.


\section{Experiments \& Results}

As we can see in Table \ref{tab:diacriticres}, with our Diacritic Detection network, the object detection approach works reasonably well. We calculate various losses to measure the performance of our model. The definition for these losses can be found in \cite{wu2017squeezedet}. Apart from the losses we are able achieve Recall as high as 0.9 with Mean Intersection Over Union (IoU) being around 0.7. The comparison results in Table \ref{tab:diacriticres} shows how our diacritic detection approach is able to outperform Squeezedet.

\begin{table}
\centering
\caption{Diacritic Detection Results}\label{tab1}
\begin{tabular}{| c | c | c |}
\hline
\bfseries{Metrics} & \bfseries{ Diacritic Detection Network} & \bfseries{Squeezedet}  \\
\hline
Class Loss &  0.31 &  3.83 \\
Bounding Box Loss &  0.09 &  0.99\\
Confidence Loss & 0.22 &  0.41 \\
Mean Intersection Over Union & 0.71 &  0.39 \\

Recall & 0.90 &  0.21 \\
\hline
\end{tabular}
\label{tab:diacriticres}
\end{table}


The next experiment which we conduct is with respect to the overall performance of the entire pipeline. We calculated multiple metrics in the form of Recall, Precision and F1 Score to have a holistic view about the performance of our pipeline. We chose 500 samples for each language from the Test dataset created as discussed in section \ref{cg}. The results in Table \ref{tab:ocrres} showcase that diacritic characters serve as an important factor even within the same script when it comes to determination of language.

\begin{table}
\centering
\caption{Language Identification Results}\label{tab1}
\begin{tabular}{| M{1.5cm} | M{1.9cm} | M{2.6cm} | M{2.6cm} | M{2.7cm} |}
\hline
\bfseries {Language} & \bfseries{Precision} & \bfseries{Recall} & \bfseries{F1 Score} \\
\hline
Spanish & 0.92 &  0.91 & 0.92  \\
German & 0.88 &  0.93 & 0.91  \\
French & 0.91 &  0.85 & 0.88  \\
Italian & 0.97 &  0.88 & 0.92  \\
Romanian & 0.95 &  0.90 & 0.93  \\
Finnish & 0.87 &  0.99 & 0.93  \\
Hungarian & 0.82 &  0.99 & 0.90  \\
Estonian & 0.98 &  0.96 & 0.97  \\
Danish & 0.95 &  0.75 & 0.84  \\
Dutch & 0.92 &  0.99 & 0.96  \\
Swedish & 0.95 &  0.71 & 0.81 \\
Portuguese & 0.75 &  0.89 & 0.82  \\
Czech & 0.90 &  0.95 & 0.92  \\
\hline
\end{tabular}
\label{tab:ocrres}
\end{table}

Apart from these results, our proposed system demonstrates efficiency with respect to device based computational restrictions. Our entire pipeline’s size is restricted to just around 5MB with inference time being as low as 213 ms. The  on-device metrics have been tabulated in Table \ref{tab:ondevice} and have been calculated  using  Samsung’s  Galaxy  A51  with  4  GB  RAM and 2.7 Ghz octa-core processor.

\begin{table}
\centering
\caption{On-Device Metrics}\label{tab1}
\begin{tabular}{| c | c | c |}
\hline
\bfseries{Component} & \bfseries{Size} & \bfseries{Inference Time} \\
\hline
Diacritic Detection Network &  5 MB &  210 ms\\
Shallow Network & 0.3 MB &  3 ms\\
Total & 5.3 MB &  213 ms\\

\hline
\end{tabular}
\label{tab:ondevice}
\end{table}

\section{Conclusion \& Future Work}

In this work, we showcase how we can identify language from text by making use of diacritic characters in images using an on-device efficient architecture with low model size and inference timings. We primarily concentrate on 13 Latin languages and observe promising results. The existing architecture can be further scaled for other Latin languages as well.

One of the areas which can be targeted as a part of future work could be to extend this work to other scripts apart from Latin. In order to achieve that, first we need to identify idiosyncratic characters in the corresponding script just like we identified diacritic characters in Latin script which can be used to differentiate between languages belonging to that script. For example in Devanagri script \footnote{https://en.wikipedia.org/wiki/Devanagari}, we have compound letters which are nothing but vowels combined with consonants. These compound letters have diacritics. Once we have diacritic or similarly identified unique set of characters, we can apply the discussed architecture and observe OCR results.

\bibliographystyle{splncs04}
\bibliography{references}

\end{document}